\documentclass{article}

\usepackage[english]{babel}

\usepackage[letterpaper,top=2cm,bottom=2cm,left=3cm,right=3cm,marginparwidth=1.75cm]{geometry}

\usepackage{amsmath}
\usepackage{graphicx}
\usepackage[colorlinks=true, allcolors=blue]{hyperref}
\usepackage{verbatim}

\title{Driving with InternVL: Oustanding Champion in the Track on Driving with Language of the Autonomous Grand Challenge at CVPR 2024}
\author{Jiahan Li, Zhiqi Li, Tong Lu \\Nanjing University}

\begin{document}
\maketitle

\begin{abstract}
This technical report describes the methods we employed for the Driving with Language track of the CVPR 2024 Autonomous Grand Challenge.
We utilized a powerful open-source multimodal model, InternVL-1.5, and conducted a full-parameter fine-tuning on the competition dataset, DriveLM-nuScenes. To effectively handle the multi-view images of nuScenes and seamlessly inherit InternVL's outstanding multimodal understanding capabilities, we formatted and concatenated the multi-view images in a specific manner. This ensured that the final model could meet the specific requirements of the competition task while leveraging InternVL's powerful image understanding capabilities. Meanwhile, we designed a simple automatic annotation strategy that converts the center points of objects in DriveLM-nuScenes into corresponding bounding boxes. As a result, our single model achieved a score of 0.6002 on the final leadboard.
\end{abstract}

\section{Introduction}
This competition primarily aimed to evaluate the perception, prediction, and planning capabilities of multimodal models in autonomous driving scenarios. Specifically, DriveLM~\cite{sima2023drivelm} designed a series of diverse natural language questions based on various autonomous driving scenarios, and the models were scored based on their responses. Different types of questions were evaluated using different scoring strategies. Notably, the competition placed greater emphasis on the perception capabilities of the multimodal models. Only if a model correctly perceives a specific object is it eligible to answer related questions.

In the following sections, we will continue to introduce the competition dataset, our methodologies, and the final results.

\section{Dataset}

DriveLM-nuScenes~\cite{sima2023drivelm,caesar2020nuscenes} consists of 378k question-answer pair in training split. As shown in Tab.~\ref{qa_example}, we show 4 examples of questions in DriveLM-nuScenes dataset.  It designed a special format to represent key objects, consisting of the object ID, camera name, and the object's center coordinates, for example \textcolor{black}{\textless c1,CAM\_BACK,1088.3,497.5\textgreater}.

We chose to change the representation of the object's center point to the object's bounding box for the following two reasons:
\begin{itemize}
    \item The representation capability of the center point is not as strong as that of the bounding box, which can provide more precise positional information about the object.
    \item Multimodal models like InternVL~\cite{chen2023internvl} inherently possess perception capabilities and can perform grounding detection or reference captioning using bounding boxes in a specific format.

In this competition, we used the Segment Anything~\cite{kirillov2023segment} model to convert object center points into object bounding boxes. Specifically, we used the object's center point as the point prompt to obtain multiple candidate masks for that point. We observed that the largest mask typically corresponds to the complete object we need. Therefore, we consistently selected the largest mask and derived the final bounding box coordinates from the mask. This method works well in most cases. However, if the object's center point is not on the main body of the object, it may produce incorrect bounding boxes. This situation can occur with traffic light objects.

\end{itemize}

\begin{table*}[htb]
\begin{center}
\begin{tabular}{|c|p{14cm}|}
\hline
 Tag&\textbf{Question}  \\
  \hline
 0& What is the moving status of object \textcolor{black}{\textless c1,CAM\_BACK,1088.3,497.5\textgreater}? Please select the correct answer from the following options: A. Going ahead. B. Stopped. C. Back up. D. Turn left. \\
 \hline
 1& What actions could the ego vehicle take based on \textcolor{black}{\textless c1,CAM\_BACK,1088.3,497.5\textgreater}? Why take this action and what's the probability? \\
  \hline
2 &\small What are the important objects in the current scene? Those objects will be considered for the future reasoning and driving decision. \\
\hline
3& What object should the ego vehicle notice first when the ego vehicle is getting to the next possible location? What is the state of the object that is first noticed by the ego vehicle and what action should the ego vehicle take? What object should the ego vehicle notice second when the ego vehicle is getting to the next possible location? What is the state of the object perceived by the ego vehicle as second and what action should the ego vehicle take? What object should the ego vehicle notice third? What is the state of the object perceived by the ego vehicle as third and what action should the ego vehicle take? \\

  \hline
\end{tabular}
\end{center}
\caption{}
\label{qa_example}
\end{table*}

\section{Model}
\begin{figure}

\centering
\includegraphics[width=0.6\linewidth]{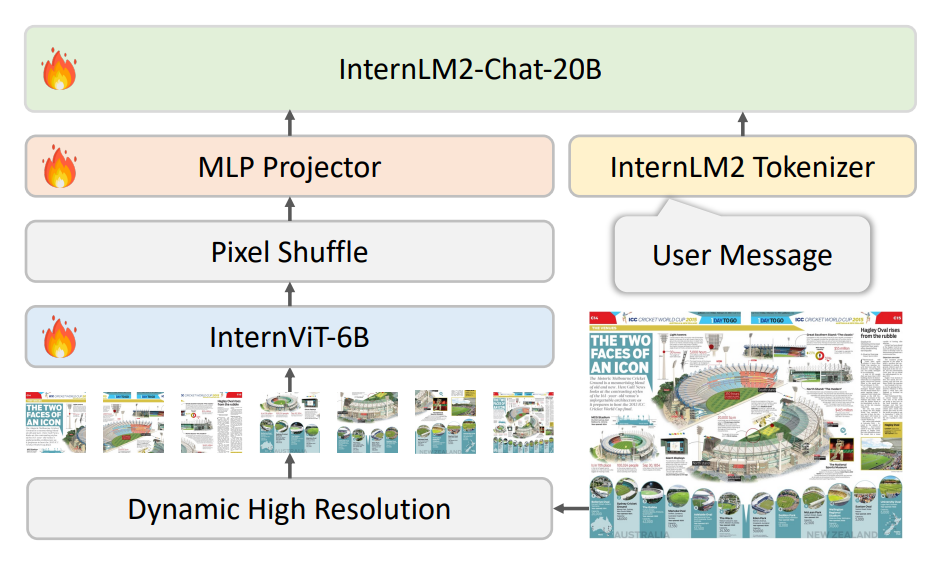}
\caption{\label{fig:internvl}Overall Architecture.}
\end{figure}

\begin{figure}
\centering
\includegraphics[width=0.7\linewidth]{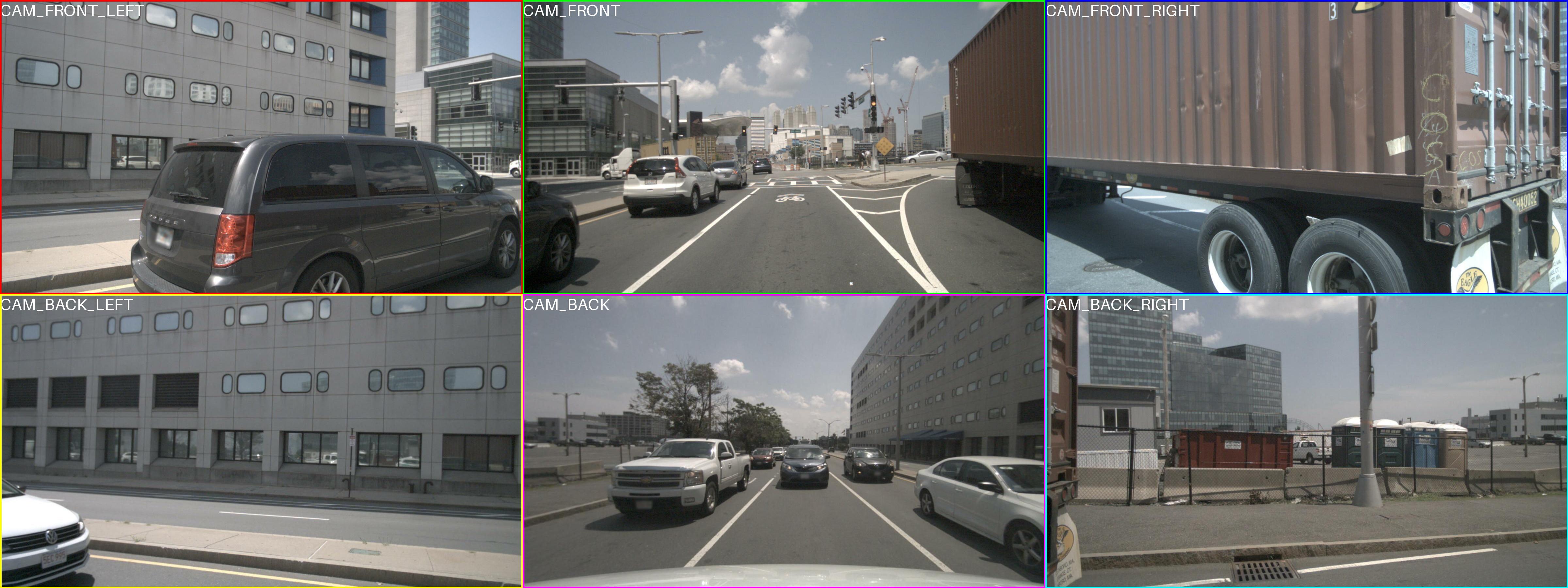}
\caption{\label{fig:whole_img}The concatenated image.}
\end{figure}

We selected InternVL-1.5 as our base model, as shown in Fig.~\ref{fig:internvl}. It consists of an InternLM-20B language model, a 6B InternViT, and a connector, and has been extensively pre-trained on multimodal data.
To handle high-resolution images, InternVL employs a dynamic high-resolution training approach that effectively adapts to the varying resolutions and aspect ratios of input images. This method leverages the flexibility of segmenting images into tiles, enhancing the model’s ability to process detailed visual information while accommodating diverse image resolutions. Although InternVL has multi-image inference capabilities, it is trained by default using a single image. Since each sample in nuScenes corresponds to six images and can also be extended temporally, we performed a concatenation operation on the multi-view images to reduce the number of images that InternVL needs to process. Specifically, we first added text to each image to indicate its orientation, such as ``CAM\_FRONT". We then resized each image to $896\times448$ pixels. The six images were arranged into a single composite image in a $2\times3$ grid. The resizing ensures easier subsequent image segmentation and preserves the integrity of each individual image as much as possible.
The final concatenated image size is 2688x896, as shown in the Figure~\ref{fig:whole_img}.

The complete image is divided into twelve 448x448 sub-images, with each view corresponding to two sub-images. Additionally, the entire image is resized to a 448x448 thumbnail for processing. Finally, each image is transformed into 256 image tokens through a VIT-MLP and pixel shuffle.

At same time,  we also include layout descriptions in the system prompt as:
\paragraph{System Prompt:} \textit{You are an Autonomous Driving AI assistant. You receive an image that consists of six surrounding camera views. The layout is as follows:
                The first row contains three images: FRONT\_LEFT, FRONT, FRONT\_RIGHT.
                The second row contains three images: BACK\_LEFT, BACK, BACK\_RIGHT.
                Your task is to analyze these images and provide insights or actions based on the visual data.}
                
It is important to note that since the large language model predicts bounding box coordinates by predicting the next token, InternVL normalizes all box coordinates to integers between 0 and 1000. Therefore, after image concatenation, we also process the bounding box coordinates accordingly to meet InternVL's requirements.

Finally, we performed full-parameter fine-tuning of InternVL-1.5 using 64 A100 GPUs. We train the model with a learning rate of 2e-5 for one epoch. We utilize the deepspeed zero-3 strategy to save memory and the batchsize is 1024.

\paragraph{Temporal Fusion} 
We also conducted preliminary explorations on temporal expansion, using the image of the previous keyframe. The corresponding input is:

\paragraph{Prompt:}\textit{System: \textless system message\textgreater\, USER: previous images: \textless image1\textgreater, current images: \textless image2\textgreater \{Question\} ASSISTANT:}

\section{Experiment}

Our experimental results are shown in Table \ref{tab:results}. Our temporal version InternVL4Drive-T had errors due to data format issues and achieved a lower score, which requires further exploration.
Our best single model InternVL4Drive-v2 achieves a final score of 0.6002. The v1 version are trained on a subset of training set, which is around 10\% of the full data. Based in this sub-dataset, our model actually achieves all higher score except on the ChatGPT score. While emplying ensemble on v1 and v2 result, actually we can obtain a much higher final score.

\begin{table*}[htbp]
\begin{center}
\resizebox{\textwidth}{!}{
\setlength{\tabcolsep}{1pt}
\begin{tabular}{lccccccccccc}
\hline
\textbf{Method} & \textbf{Accuracy$\uparrow$} & \textbf{ChatGPT$\uparrow$} & \textbf{Bleu\_1} & \textbf{Bleu\_2} & \textbf{Bleu\_3} & \textbf{Bleu\_4} & \textbf{ROUGE\_L} & \textbf{CIDEr} & \textbf{Match$\uparrow$} & \textbf{Final Score$\uparrow$} \\
\hline

InternVL4Drive-v1 & {0.7718} & {59.9800} & {0.7940} & {0.7317} & {0.6741} & {0.6185} & {0.7463} & {0.2100} & {47.9204} & {0.5862} \\
InternVL4Drive-v2 & {0.7339} & {65.2512} & {0.7787} & {0.7176} & {0.6608} & {0.6059} & {0.7449} & {0.2061} & {47.6482} & {0.6002} \\
InternVL4Drive-T & {0.2080} & {61.1232} & {0.7091} & {0.6505} & {0.5957} & {0.5428} & {0.7257} & {0.1768} & {41.2762} & {0.4600} \\
\hline
\end{tabular}}
\end{center}
\caption{The results on DriveLM dataset.}
\label{tab:results}
\end{table*}

\bibliographystyle{alpha}
\bibliography{sample}

\end{document}